\documentclass[lettersize,journal]{IEEEtran}
\usepackage{amsmath,amsfonts}
\usepackage{algorithmic}
\usepackage{algorithm}
\usepackage{array}
\usepackage[caption=false,font=normalsize,labelfont=sf,textfont=sf]{subfig}
\usepackage{textcomp}
\usepackage{stfloats}
\usepackage{url}
\usepackage{verbatim}
\usepackage{graphicx}
\usepackage{cite}
\usepackage{amsmath,amsfonts,bm}

\def\eqref#1{equation~\ref{#1}}

\def\1{\bm{1}}

\def\va{{\bm{a}}}
\def\vb{{\bm{b}}}

\def\vr{{\bm{r}}}
\def\vs{{\bm{s}}}

\def\vx{{\bm{x}}}

\def\vz{{\bm{z}}}

\def\mA{{\bm{A}}}

\def\mR{{\bm{R}}}
\def\mS{{\bm{S}}}

\DeclareMathAlphabet{\mathsfit}{\encodingdefault}{\sfdefault}{m}{sl}
\SetMathAlphabet{\mathsfit}{bold}{\encodingdefault}{\sfdefault}{bx}{n}

\newcommand{\R}{\mathbb{R}}

\DeclareMathOperator*{\argmax}{arg\,max}

\usepackage{multirow}
\usepackage{xcolor}

\begin{document}

\title{TeViR: Text-to-Video Reward with Diffusion Models for Efficient Reinforcement Learning\thanks{This work has been submitted to the IEEE for possible publication. Copyright may be transferred without notice, after which this version may no longer be accessible.}}

\author{Yuhui Chen, 
Haoran Li,
Zhennan Jiang,
Haowei Wen,
Dongbin Zhao
\thanks{Yuhui Chen, Haoran Li, Zhennan Jiang, and Dongbin Zhao are with the Institute of Automation, Chinese Academy of Sciences, Beijing 100190, China, and are also with the School of Artificial Intelligence, University of Chinese Academy of Sciences, Beijing 100049.}
\thanks{This work was done while Haowei Wen was a research intern at Institute of Automation, Chinese Academy of Sciences, Beijing 100190, China.}
}



\maketitle

\begin{abstract}
Developing scalable and generalizable reward engineering for reinforcement learning (RL) is crucial for creating general-purpose agents, especially in the challenging domain of robotic manipulation. While recent advances in reward engineering with Vision-Language Models (VLMs) have shown promise, their sparse reward nature significantly limits sample efficiency. This paper introduces TeViR, a novel method that leverages a pre-trained text-to-video diffusion model to generate dense rewards by comparing the predicted image sequence with current observations. Experimental results across 13 simulation and real-world robotic tasks demonstrate that TeViR outperforms traditional methods leveraging sparse rewards and other state-of-the-art (SOTA) methods, achieving better sample efficiency and performance without ground truth environmental rewards. TeViR's ability to efficiently guide agents in complex environments highlights its potential to advance reinforcement learning applications in robotic manipulation.
\end{abstract}


\section{Introduction}
\IEEEPARstart{D}{eveloping} general-purpose agents with reinforcement learning (RL) necessitates scalable and generalizable reward engineering to provide effective task specifications for downstream policy learning \cite{ma2023towards}. Reward engineering is crucial as it determines the policies agents can learn and ensures they align with intended objectives. However, the manual design of reward functions often present significant challenges \cite{he2021reinforcement, chai2023hierarchical, chen2025conrft}, particularly in robotic manipulation tasks \cite{avi2019end, karoly2021deep, li2020robot, minae2023reward}. This challenge has emerged as a major bottleneck in developing general-purpose agents.

Although inverse reinforcement learning (IRL) \cite{saurabh2021survey} learns rewards from pre-collected expert demonstration, these learned reward functions are unreliable for learning policies due to noise and misspecification errors \cite{Dario2016concrete}, especially for robotic manipulation tasks since in-domain data is limited \cite{yecheng2023vip}. Additionally, the learned reward functions is not generally applicable across tasks. With the impressive generalization capabilities, pre-trained Vision-Language Models (VLMs) enables referencing visual features with natural language \cite{alec2021learning}, and thus used to generate sparse rewards for policy training by calculating the similarity between language descriptions of tasks and the visual observations \cite{jingyun2023robot, sumedh2023roboclip}. However, the reward signals generated in these methods are often unstable and meets high variance, leading to insufficient supervision and low sample efficiency. This is because VLMs lack the diversity to generalize to different tasks as they are trained only on matching language descriptions and visual observations \cite{yufei2024reinforcement}.

Recent approaches about pre-trained video diffusion models \cite{ho2022video} have achieved significant success in capturing the complex distribution of videos, such as the text-to-video generation \cite{ho2022imagen}. This technology is then employed in policy modeling, using text-to-video diffusion models to predict future frames and inverse dynamics models to derive actions for decision-making \cite{yilun2023learning, chen2023learning, mengjiao2023learning, chuan2024any}. However, this require high-quality video prediction \cite{ye2023foundation} or otherwise the inverse dynamics model may fail to generate appropriate actions. Additionally, some methods generate video prediction rewards through prediction likelihood \cite{alejandro2023video, tao2023diffusion} to encourage the robot to match the desired behavior. However, these methods often lead to sub-optimal behaviors since the prediction likelihood only generates the short-term image under the policy \cite{haoran2025stabilizing}. These approaches minimize prediction uncertainty rather than providing effective guidance for achieving long-term goals. More importantly, these methods still rely on the sparse reward from the environment.

To this end, we present \textbf{Te}xt-to-\textbf{Vi}deo \textbf{R}eward (TeViR), a novel framework designed to calculate dense rewards by generating videos with the language instruction input, enabling agents to efficiently learn the desired policy. Our framework involves video generation for planning the task execution and assessing whether current observation meet expectations and contribute to task progress. Visual and linguistic information play crucial roles in human learning and decision-making processes \cite{padilla2018decision, yilun2023learning, li2024mat}. We combine these two types of information with prior information across different tasks, and utilizes text-to-video diffusion models for long horizon planning, which generates observations of the imaginary trajectory based on the current visual observation and language task description. By comparing these predicted images with the current observations, dense rewards can be calculated for each timestep, without relying on the environment rewards. Such dense rewards enable the policy to evaluate the current state and efficiently learn the desired behavior \cite{yuhui2024boosting, gu2022proximal}. Benefits from the long-term planning ability of the video generation model, TeViR exploits the information from generated trajectory, thereby achieving better performance and sample efficiency compared to predicting the short-term observations.

Concretely, this work makes the following contributions:
\begin{enumerate}
    \item We propose TeViR, a novel framework that annotates reward for RL agents. By utilizing the capability of text-to-video diffusion models to generate videos, TeViR relies solely on the text description of the task and the visual observations to generate dense rewards for policy learning, eliminating the dependency on the ground truth environmental reward.
    \item TeViR offers a multi-view approach for video generation to tackle occlusion issues. It calculates dense rewards by assessing the distance to the desired trajectory and the progress toward the goal. This reward system enhances the sample efficiency and success rate of RL policies. It can be applied to learn policies for various robotic manipulation skills, such as reaching, pushing, and grasping.
    \item We conduct experiments to show that TeViR achieves the new SOTA for the reward engineering in 11 Meta-World tasks and 2 real-world tasks. Notably, without the environmental rewards, the average success rate increased by 49.1\% on Meta-World tasks and by 29.0\% on real-world tasks within same training steps.
\end{enumerate}

Through these contributions, our approach addresses the limitations of existing methods and offers a robust framework for reward engineering in complex RL environments.

\begin{figure*}[t]
\centering
\includegraphics[width=\linewidth]{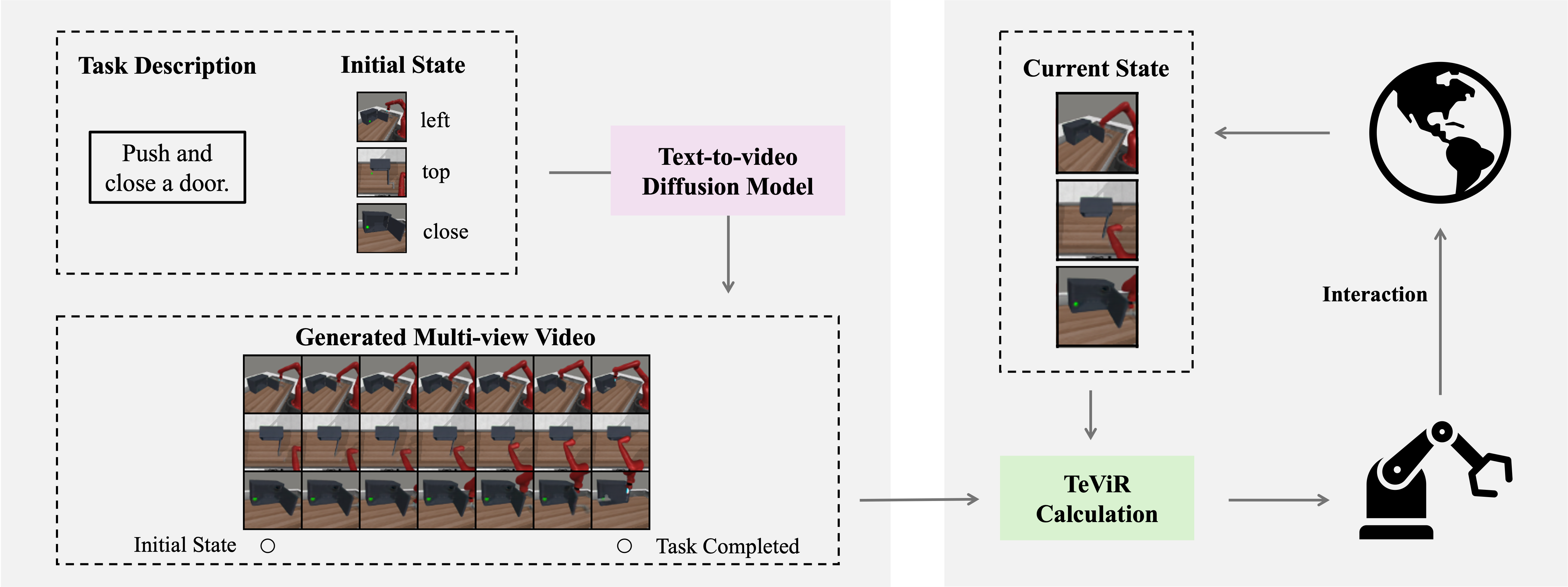}
\caption{Overall framework of TeViR. The text-to-video diffusion model takes the initial RGB observation of the environment and a textual goal description as its input. Then it synthesizes an image sequence of the future and the TeViR is calculated using the current observation and the synthesized image sequence. Finally, the TeViR helps the agent update its policy.}
\label{fig:overview}
\end{figure*}

\section{Related Work}
\subsection{Learning from Expert Videos}
How to learn policy from expert videos has been studied for decades. Imitation learning (IL) methods often rely on state-action pairs from expert demonstrations to learn policy. Recent studies such as \cite{corey2022interactive, jyothish2022the, yang2023watch}, emphasize learning directly from the raw export data, bypassing the need for manually labeled data. IRL \cite{pieter2004apprenticeship, chelsea2016guided, kevin2021xirl} extends the capabilities of IL by deriving reward functions from the observed behaviors of experts. Several works use perceptual metrics such as goal-image similarity and trajectory similarity to calculate rewards \cite{pierre2017unsupervised, annie2021learning, yecheng2023vip, jialu2023can}. A novel approach involves using video prediction models to estimate future states and derive rewards based on predicted log-likelihood or entropy \cite{jingyun2023robot, alejandro2023video, tao2023diffusion}. These methods allow for dynamic reward generation, potentially increasing the adaptability and efficiency of learning algorithms. In contrast to prior methods, our approach utilizes a text-to-video diffusion model to generate dense rewards by comparing predicted video sequences with the agent's current observations, providing a more scalable and generalizable feedback mechanism for RL.

\subsection{Large models as Reward Functions}
To design reward functions capturing human preferences, methods \cite{minae2023reward, hengyuan2023language} explores large language models (LLMs) in text-based games like negotiations or card games. Other works followed by demonstrating that LLMs can write code of the reward function for training robots \cite{wenhao2023language, yufei2023robogen, yecheng2023euraka}. Focusing on applications that interact with the physical world, approaches employ VLMs as a success detector to provide sparse task rewards \cite{yuqing2023vision}, or to evaluate whether image observation of agent aligns with language description of the robot tasks \cite{yuqing2023guiding, sumedh2023roboclip, juan2023vision, ma2023liv, chang2024large}. However, these methods operate on individual frames or short clips, without modeling full temporal task completion, and their reward accuracy heavily depends on how well the agent’s visual observations align with the data distribution of the pre-trained VLMs\cite{sumedh2023roboclip, yecheng2023vip}. This limits their effectiveness in long-horizon, open-ended tasks, particularly when domain or viewpoint shifts occur.

Other than using language as the interface, more recent approaches applied the large Text-to-Image model to generate visual goal \cite{jialu2023can} or web-scale text-to-video diffusion model to learn universal policies \cite{yilun2023learning, chen2023learning, kevin2023zero}, serving image as a unified representation of the physical world. In contrast, our method learns dense rewards via a text-to-video diffusion model conditioned on both language descriptions and current visual observations, offering more informative and continuous feedback to accelerate policy learning while improving the scalability and generalization across various tasks.

\begin{figure*}[t]
\centering
\includegraphics[width=\linewidth]{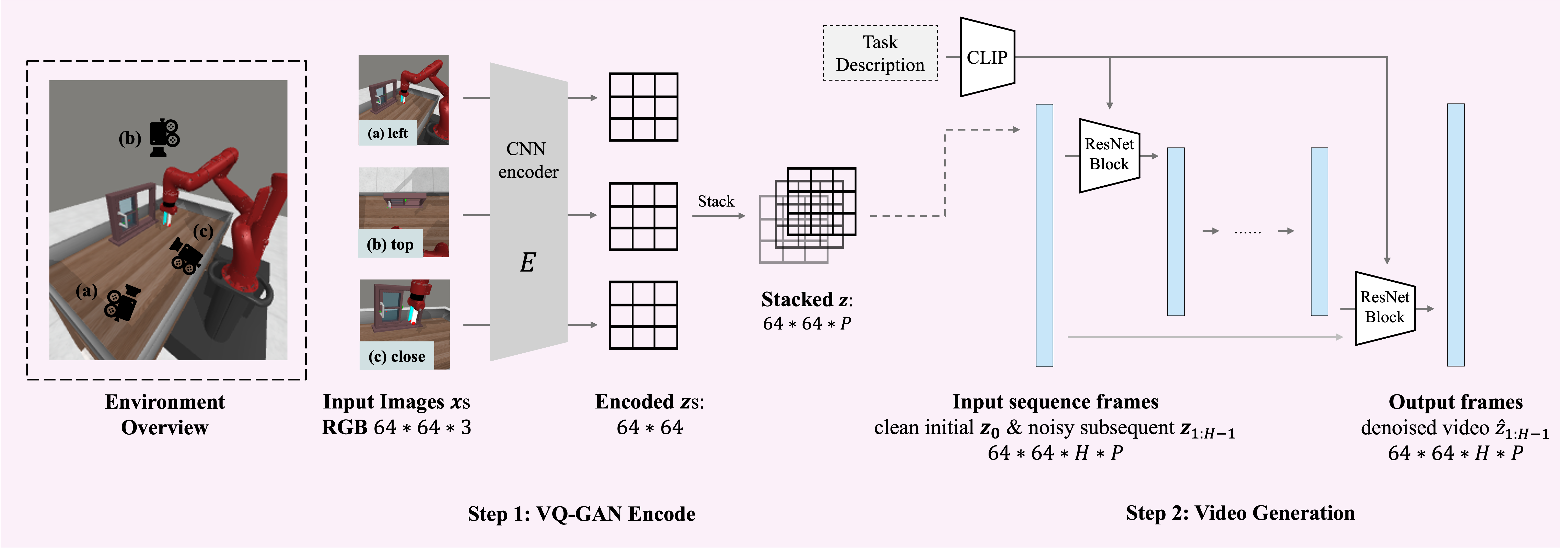}
\caption{Illustration of the 3 view selected for the Meta-World environment and the network architecture of our text-to-video diffusion model. We chose $left$, $top$ and $close$ to collect visual observations, so as to comprehensively capture the robot state during the task execution. (a),(b) and (c) are the images observed from 3 view, respectively. The video generation process includes 2 steps: (1) We use an CNN-based encoder to encode RGB images into vectors $\vz$ (2) Then we use an U-Net architecture with factorized spatial-temporal convolution kernel as the basic building block for video generation.}
\label{fig:vdm}
\end{figure*}

\section{Preliminaries}
\subsection{Markov Decision Process}
We consider an RL agent that interacts with the environment modeled as a Markov Decision Process (MDP) with a finite horizon. The process is defined by a tuple $\{\mS, \mA, P, \mR, \gamma\}$, where $\mS$ and $\mA$ represents the state space and the action space. For action $\va_t \in \mA$, state and next state $\vs_t , \vs_{t+1} \in \mS$ at time step $t$, $P(\vs_{t+1}|\vs_t,\va_t)$ represents the environment transition function, $r(\vs_t,\va_t,\vs_{t+1}) \in \mR$ represents the reward function, and $\gamma$ represents the discount factor. The goal of this RL agent is to learn an optimal policy $\pi(\va_t|\vs_t)$ that maximizes the expected return $\mathbb{E}_{t=0}^{T-1}[\gamma^tr_t]$. And we use subscripts $t\in \{0,\cdots,T-1\}$ to denote trajectory time steps of a finite-horizon $T$. 

\subsection{Text-to-video Diffusion Model}
The text-to-video diffusion model is a conditional video diffusion model \cite{song2020denoising, ho2022video} that takes the initial image $\vx_0$ and a language description $txt$ as its condition and aims to approximate the distribution $p_{\phi}(\bm{\tau}_{1:H-1}|\vx_0, txt)$, where $\bm{\tau}_{1:H-1} = \{\vx_1, ..., \vx_{H-1}\}$ represents a video clip from video time step $1$ to $H-1$.

The forward diffusion process applies noise $\bm{\epsilon}$ in the latent space at each diffusion time step $k \in \{ 0, ..., K\}$ to the data distribution $\bm{\tau}_{1:H-1}$. The noisy sample can be calculated by $\bm{\tau}_{1:H-1}^k = \sqrt{\overline{a_t}}\bm{\tau}_{1:H-1}^0 + \sqrt{1-\overline{a_t}}\epsilon$, where  $\overline{a_t}$ is the accumulation of the noise schedule over past time steps $\{0,..., k\}$. We use superscripts $k\in \{0,\cdots,K-1\}$ to denote diffution time steps. To learn the distribution $p(\bm{\tau}_{1:N-1}|\vx_0, txt)$, this diffusion model aims to train a score function $\epsilon_\theta$ to predict the noise $\bm{\epsilon}$ that applied to $\bm{\tau}_{1:N-1}$ given the perturbed sample. Given the Gaussian noise scheduling $\overline{a_t}$, the overall learning objective with mean squared error (MSE) is shown below:
\begin{equation}
    \begin{aligned}
    L_{MSE} &= ||\bm{\epsilon} - \epsilon_\theta(\bm{\tau}_{1:H-1}^k|\vx_0, txt)||^2 \\
    &= ||\bm{\epsilon} - \epsilon_\theta(\sqrt{\overline{a_t}}\bm{\tau}_{1:H-1}^0 + \sqrt{1-\overline{a_t}}\epsilon|\vx_0, txt)||^2 \\
    \end{aligned}
\label{eq:diffusion loss}
\end{equation}
where the noise $\bm{\epsilon}$ is sampled from a multivariate Gaussian distribution, and $k \in \{ 1, ..., K\}$ is a randomly sampled diffusion step.

\section{Method}
\subsection{TeViR Framework}
To generalize TeViR across different tasks and environments, we consider an RGB images $\vs \in \R^{\mathcal{H}\times \mathcal{W}\times 3}$, as a universal interface between the RL agent and environments. However, reward engineering on RGB images without reliance on the environmental sparse reward is tedious and sometimes intractable. To tackle this prominent challenge, we propose our framework for dense reward calculation which consists of three modules, a text-to-video diffusion model to synthesize expert future image sequences conditioned on the first frame and the task descriptions, the TeViR to calculate rewards, and an RL agent using the reward to update its policy. 

As depicted in Figure \ref{fig:overview}, our framework leverages expert experience from pre-trained text-to-video diffusion models at a high level. The model takes a language description to specify desired task and an RGB image to indicate the initial state of the environment as condition, and generates a sequence of key frames illustrating the expected progression toward task completion. We expect the RGB image to capture dynamic features in the environment, with the generated expert video bridging the domain gap between task language description and the robot behavior. This enables more comprehensive guidance for long-horizon tasks, addressing the challenges posed by sparse or unreliable reward signals.

Downstream, to adapt the RL agent to the target behavior represented under the generated video, our framework calculates dense rewards using the generated video and current observation to encourage it exploring and learning the desired policy. In contrast to existing methods like RoboCLIP \cite{sumedh2023roboclip}, which provide sparse rewards based on task language and visual observation similarity, and Diffusion Reward \cite{tao2023diffusion}, which only predicts short-term images without capturing long-term task progression, our framework generates dense rewards by comparing the current observation to the entire predicted image trajectory. This allows the RL agent to receive continuous, informative feedback throughout the task, rather than relying on a single sparse signal. By calculating rewards based on both visual observation and the predicted image sequence, our framework isolates the reward calculation process from any environmental information other than the visual inputs, while still leveraging prior knowledge learned from the text-to-video diffusion models. This unique design leads to more effective exploration and better learning of the desired policy in complex robotic tasks.

\subsection{Multi-view Video Generation}
We then illustrate how we generate videos through a pre-trained text-to-video model. Firstly, we train an image encoder to condense the information from the high-dimensional observation space. Specifically, we employ the VQ-GAN\footnote{\url{https://github.com/dome272/VQGAN-pytorch}} encoder \cite{patrick2021taming} to encode the image $\vx$ into a quantized latent vector $\vz = Q(E(\vx))$, where $E$ represents the encoder and $Q$ denotes the element-wise quantizer. With this encoder, an image $\vx$ is then represented by a latent vector $\vz$ in the following sections.

\begin{figure*}[t]
\centering
\includegraphics[width=0.9\linewidth]{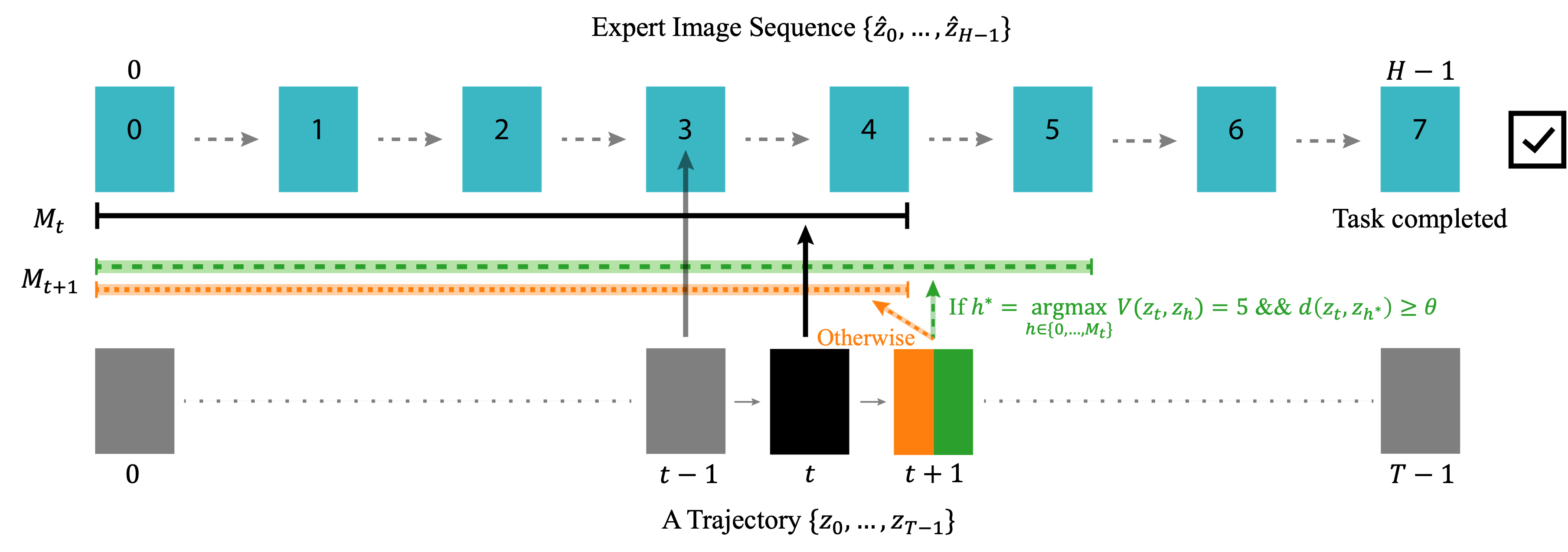}
\caption{Progress calculation illustration. For each image $z_t$ in a trajectory, the progress reward calculates the similarity with the corresponding generated image sequence within its progress $M_t$. We assume that the image $z_{t-1}$ on a trajectory with progress $M_{t-1}=4$ is similar to the farthest reached image $z_{M_{t-1}}$ and we have $M_t=5$. If $z_t$ of the trajectory is similar to image $z_{M_{t}}$, then progress for $z_{t+1}$ should be $6$, labeled in green. Otherwise, the progress for $z_{t+1}$ would maintain $5$, labeled in orange.}
\label{fig:reward}
\end{figure*}

Since RGB images capture only color information, they have limitations in dealing with occlusions and offering a comprehensive spatial understanding of the environment. To address these challenges, we utilize multiple $P$ views, which provide complementary perspectives that enhance the model's ability to assess the mechanical structure and task execution. The selection of these views is made based on their ability to capture different aspects of the task and may be adjusted according to specific environment and task requirements. For the robot manipulation tasks we selected, we choose $P=3$ views: ${left, top, close}$, as illustrated on the left in Figure \ref{fig:vdm}. The left view ($\vz^{left}$) offers a broad, global perspective of the environment, ensuring that the agent has an overall understanding of the task setting. The top view ($\vz^{top}$) provides a vertical perspective, which is essential for tasks that require precise positioning or spatial awareness from above, such as aligning with objects. The close view ($\vz^{close}$) focuses on the robot and task object, offering fine-grained details of the manipulation process, such as grasping or object interaction. These diverse views help address occlusion issues and ensure that the model captures a complete spatial representation. To simplify the notation, if a vector $\vz$ is referenced later without a superscript indicating a specific view, it should be interpreted as representing the aggregation of all selected views.

The whole architecture of our text-to-video diffusion model is illustrated in Figure \ref{fig:vdm}. Firstly, the images from all view are firstly encoded into latent vectors the VQ-GAN encoder. Then we stack all latent vectors as one frame, a basic block of the sequence frame for video prediction via an U-Net model. The U-Net model finally takes a clean initial frame $\vz_0$ and $H-1$ subsequent noisy frames $\vz_{1:H-1}$ as input and generate the image sequence $\widehat{\vz}_{0:H-1}$. By combining latent vectors from all views, we create a more complete and detailed spatial representation of the scene. Also, stacking latent vectors ensures that information from different views is integrated into a single coherent frame, maintaining consistency and reducing discrepancies that may arise from individual perspectives. Throughout this paper, the text-to-video diffusion model predicts a fixed number of image sequence, where we set $H = 8$ in all experiments. 

\begin{figure*}[ht]
\centering
\includegraphics[width=\linewidth]{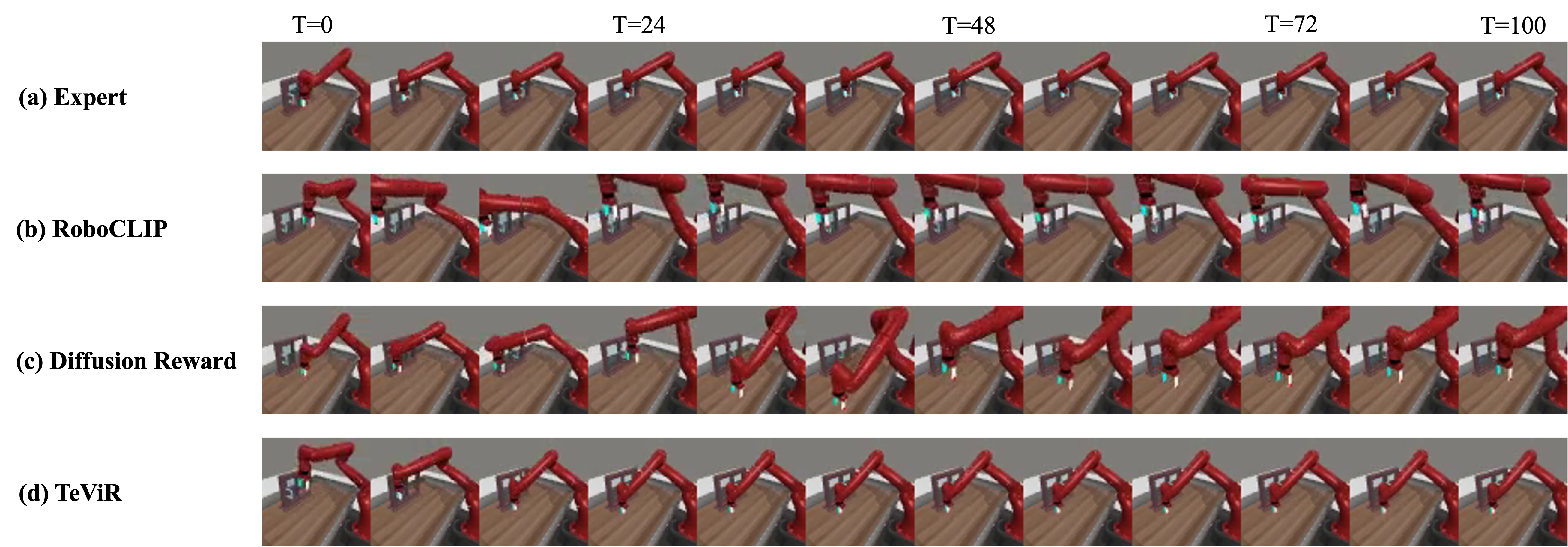}
\caption{Trajectory visualizations of 'window-close-v2' from $left$ view. (a) The scripted policy provided by Meta-World demonstrates expert performance. (b) The policy trained by RoboCLIP approaches the target object, the window, but fails to close it. (c) The policy trained by Diffusion Reward learns to close the window but exhibits instability, resulting in a low success rate. (d) The policy trained by TeViR successfully learns the desired behavior with high robustness.}
\label{fig:traj}
\end{figure*}

\subsection{Text-to-video Reward}
Finally, we introduce the details of our reward calculation method. The reward functions is important to guide the agent's behavior by indicating the desirability of its actions to maximize long-term success. Therefore, the critical insight behind TeViR is to how closely the current observation aligns with the generated image sequence and the overall progress toward completing the task. This evaluation allows us to calculate a reward that effectively encourages the agent to prioritize states and actions that lead to the desired trajectory, ultimately reinforcing behaviors that align with successful task completion.

In designing the TeViR, we aim to capture the essence of effective learning by rewarding observations that closely resemble the final stages of the generated image sequence. This strategic alignment encourages the agent to prioritize states that are nearer to the desired trajectory, thus reinforcing behaviors that closely mirror expert performance. By focusing on these pivotal moments, the TeViR effectively guides the agent towards the goal, ensuring that it hones in on the most crucial aspects of the task. 

By interacting with the environment for a whole rollout, we have a trajectory $\{\vz_0, ..., \vz_{T-1}\}$ and a generated image sequence $\{\widehat{\vz}_0, ..., \widehat{\vz}_{H-1}\}$. The trajectories are evaluated based on these main criteria:

\paragraph{Distance Reward}
To guide the agent towards states that closely match the desired trajectory, we firstly introduce the distance reward $r^{dist}_t$. A naturalistic way to define the distance between 2 states is through similarity. We define a function to quantify the similarity between corresponding images on the trajectory $\vz_{1:T-1}$ and the generated image sequence $\widehat{\vz}_{1:H-1}$. This function captures the feature-based discrepancies between the two trajectories. Since the agent’s observations are derived from multiple views (e.g., left, top, and close-up), the reward calculation must take into account the information from all these perspectives. Specifically, we calculate the cosine similarity $\sigma(\va, \vb) = \frac{\va\cdot \vb}{||\va|| \cdot ||\vb||}$ for each view and combine the results using a weighted average across different views:
\begin{equation}
    \begin{aligned}
    \sigma(\vz_t, \widehat{\vz}_h) &= \frac{\sum_{i=0}^{P}w^{view_i}\sigma(\vz_t^{view_i}, \widehat{\vz}_h^{view_i})}{\sum_{i=0}^{P}w^{view_i}}\\
    \end{aligned}
\label{eq:dis}
\end{equation}
where $w^{view_i}, p \in \{1, ..., P\}$ are hyperparameters for different view to adapt to different task requirements. This ensures that different perspectives contribute appropriately to the reward based on the task’s requirements. We give the definition of the distance reward:
\begin{equation}
    \begin{aligned}
    \vr^{dist}_t &= \sigma(\vz_t, \widehat{\vz}_{h^*})\\
    \end{aligned}
\end{equation}
where $h^* = \mathop{\argmax}_{h \in \{ 0, ... M_t-1\}} \sigma(\vz_t, \widehat{\vz}_h)$.

\paragraph{Progress Reward}
We then have the progress reward $r^{prog}_t$, for evaluating the overall progress toward completing the task. Consider the task of opening a door. If the agent skips the step of grasping the doorknob and tries to pull the door directly, it fails to complete the task correctly. Therefore, we should guide the agent to perform each critical step in sequence to ensure the proper execution of the task. 

To mitigate the agent's propensity to skip essential intermediate steps and jump directly to the end of the trajectory, we introduce a concept termed a reached image. This mechanism employs a similarity threshold $\theta$ to evaluate the progress $M$ that the agent has made towards replicating the expert image sequence, where $M_t \in \{0, ..., H-1\}$ that marks the farthest image along the expert image sequence that the agent has successfully approximated up to trajectory time step $t$. This mechanism is illustrated in Figure \ref{fig:reward}. 

Then, a binary reward is issued if the agent's current observation closely resembles the expert's final image in the trajectory. When the environment sparse reward $\vr^{spar}$ is available in some cases, we can replace the corresponding part with the ground truth value directly and refer as $\vr^{prog+}_t$. We give the definition of the progress reward $r^{prog}_t$:
\begin{equation}
    \begin{aligned}
    &\vr^{prog}_t = \alpha h^* + \mathbb{I}[\sigma(z_t, \widehat{\vz}_{H-1})>\theta]\\
    &\vr^{prog+}_t = \alpha h^* + r^{spar}_t\\
    \end{aligned}
\end{equation}
where $h^* = \mathop{\argmax}_{h \in \{ 0, ... M_t-1\}} \sigma(\vz_t, \widehat{\vz}_h)$ and $\alpha$ is a hyperparameter for balancing the progress reward. In practice we find $\alpha=0.125$ is suitable for all tasks selected.

\begin{algorithm}[t]
    \caption{TeViR Formulation} 
    \begin{algorithmic}
        \STATE \textcolor{gray!90}{\# Pretrain a text-to-video diffusion model $\phi$ if not available}
        \STATE Collect expert videos and corresponding task description $\mathcal{D}$ from multiple tasks
        \STATE Train diffusion model $p_{\phi}$ on dataset $\mathcal{D}$ by Equation (\ref{eq:diffusion loss})
        \STATE \textcolor{gray!90}{\# Downstream RL with TeViR}
        \FOR{each rollout}
            \STATE Initialize reached image $M_0=0$
            \STATE Observe $\vz_0$ and generate the image sequence $\{\widehat{\vz}_0, ..., \widehat{\vz}_{H-1}\}$
            \STATE Collect a trajectory $\{\vz_0, ..., \vz_{T-1}\}$
            \FOR{$t \in \{0, 1, ..., T-1\}$}
                \STATE Calculate the TeViR or TeViR+ by Equation (\ref{eq:reward_nenv}), Equation (\ref{eq:reward_env})
                \IF{Current image is similar to the farthest reached image: $\sigma(\vz_t, \widehat{\vz}_{M_t-1}) \geq \theta$} 
                \STATE $M_{t+1} = M_{t}+1$ \textcolor{gray!90}{\# Update the reached image}
                \ELSE 
                \STATE $M_{t+1} = M_{t}$ \textcolor{gray!90}{\# Keep the reached image}
                \ENDIF
            \ENDFOR
            
        \ENDFOR
    \end{algorithmic} 
    \label{alg:tevir}
\end{algorithm}

\paragraph{Exploration reward}
While the above-mentioned rewards encourage the agent to replicate expert behaviors, the complexity of tasks with high-dimensional inputs may still limit exploration. To address this, we integrate Random Network Distillation\footnote{\url{https://github.com/jcwleo/ random-network-distillation-pytorch}} \cite{yuri2018exploration} as an exploration reward, referred to as $\vr^{expl}_t = \vr^{rnd}_t$. It allows the agent to quickly learn how to recover from suboptimal or undesirable states, thereby improving sample efficiency.

The $\vr^{rnd}_t$ is a reward calculation method that encourages exploration by introducing an intrinsic reward based on prediction errors. Specifically, it uses a fixed randomly initialized neural network $\nu_0$ and a predictor network $\nu_1$. The agent receives a higher exploration reward when the predictor network $\nu_1$ struggles to accurately predict the output of the random network $\nu_0$ given the same input. This incentivizes the agent to explore states that are novel or less familiar, as these states typically yield higher prediction errors and thus higher exploration rewards.

We finally give the definition of the TeViR:
\begin{equation}
    \begin{aligned}
    r^{TeViR}_t &= r^{dist}_t + r^{prog}_t + r^{expl}_t\\
    \end{aligned}
\label{eq:reward_nenv}
\end{equation}
When the environment sparse reward $\vr^{spar}$ is available, we can have:
\begin{equation}
    \begin{aligned}
    r^{TeViR+}_t &= r^{dist}_t + r^{prog+}_t + r^{expl}_t\\
    \end{aligned}
\label{eq:reward_env}
\end{equation}
We summarize the complete reward formulation procedure of TeViR in Algorithm \ref{alg:tevir}.

\section{Experiments}
Our experiments are conducted in a simulated environment that reflects the complexities and sparse nature of real-world robotic manipulation scenarios. The effectiveness of our reward function is evaluated against traditional reward strategies, focusing on the agent's ability to learn correct behaviors efficiently. All agents are trained using raw pixel observations from the environment without access to other information, such as state or task rewards. In our experiments, we aim to answer the following questions:

\begin{enumerate}
    \item Is TeViR capable of efficiently guiding an agent to learn the desired behavior rapidly? (Section \ref{sec:exp_nenv})
    \item Is TeViR effective at providing a sufficient learning signal for learning the desired behavior without environmental rewards? (Section \ref{sec:exp_env})
    \item Can TeViR maintain robustness despite poor text-to-video generation quality? (Section \ref{sec:exp_unipi})
    \item What implementation detail of TeViR matters for task learning?  (Section \ref{sec:abla})
\end{enumerate}

\subsection{Experimental Setup}
\subsubsection{Baselines}
We compare our method against the following methods:

\begin{itemize}
    \item \textbf{Raw Sparse Reward}: This method only uses the ground truth environment sparse reward for training the policy. This is a basic policy training reward for traditional RL methods.
    
    \item \textbf{RoboCLIP}: RoboCLIP\footnote{\url{https://github.com/sumedh7/RoboCLIP}} \cite{sumedh2023roboclip} utilizes pre-trained VLMs to generate rewards for the agent. This is done by providing a sparse reward to the agent at the end of the trajectory which describes the similarity of the agent's behavior to the language task description. This comparison tests the benefit of generating dense rewards for policy learning.

    \item \textbf{VIPER}: VIPER\footnote{\url{https://github.com/Alescontrela/viper_rl}} \cite{alejandro2023video} learns a video prediction transformer from expert videos and directly leverages the video model's likelihood as a reward signal to encourage the agent to match the video model's trajectory distribution. This comparison tries to answer the benefit of using text-to-video diffusion models to generate the whole expert trajectory for reward formulation.

    \item \textbf{Diffusion Reward}: Instead of VIPER using a video prediction transformer, Diffusion Reward\footnote{\url{https://github.com/TEA-Lab/diffusion_reward}} \cite{tao2023diffusion} leverages conditional video diffusion models to capture the expert video distribution. Then, it utilizes the conditional entropy of the video prediction with the ground truth environment sparse reward to formulate dense rewards. This analysis aims to evaluate the advantages of the guidance ability of TeViR for efficient policy convergence. To compare the method under a situation where an environment reward is not available, we further implement the Diffusion Reward without the ground truth environment sparse reward, labeled as Diffusion Reward- in the following section.

    \item \textbf{UniPi}: Unlike other reward engineering methods, UniPi\footnote{\url{https://github. com/flow-diffusion/AVDC}} \cite{yilun2023learning} learns a text-to-video diffusion model from internet-scale video datasets as a planner, and then an inverse dynamics model is used to extract the underlying actions from the generated video. This comparison evaluates the robustness of our method in terms of the generation quality of the video. For the inverse dynamic model of UniPi, we follow the network architecture described in Appendix A.2 of UniPi \cite{yilun2023learning}, which consists of a 3x3 convolutional layer, 3 layers of 3x3 convolutions with residual connection, a mean-pooling layer across all pixel locations, and an MLP layer of (128, 4) channels to predict the action for robot control.
\end{itemize}

\begin{table}[ht]
  \begin{center}
    \caption{Language description and hyperparameters of TeViR for each task}
    \label{tab:param}
    \resizebox{\columnwidth}{!}{
    \begin{tabular}{c|ccc}
        \textbf{Task}           &\textbf{Language Description}      &$\theta$ &\{$w^{left}, w^{top}, w^{close}$\} \\
        \hline
        button-press-topdown-v2 &Press a button from the top.         &0.8      &\{0.5, 0.8, 0.4\} \\
        button-press-v2         &Press a button.                      &0.8      &\{0.5, 0.8, 0.4\} \\
        coffee-button-v2        &Push a button on the coffee machine. &0.8      &\{0.4, 0.8, 0.5\} \\
        door-close-v2           &Close a door.                        &0.9      &\{0.2, 0.8, 0.3\} \\
        door-open-v2            &Open a door.                         &0.7      &\{0.3, 0.8, 0.5\} \\
        drawer-close-v2         &Close a drawer.                      &0.8      &\{0.8, 0.4, 0.4\} \\
        drawer-open-v2          &Open a drawer.                       &0.8      &\{0.5, 0.2, 0.7\} \\
        handle-press-side-v2    &Press a handle down sideways.        &0.8      &\{0.3, 0.8, 0.5\} \\
        handle-press-v2         &Press a handle down.                 &0.8      &\{0.3, 0.8, 0.5\} \\
        window-close-v2         &Close a window.                      &0.8      &\{0.2, 0.8, 0.2\} \\
        window-open-v2          &Open a window.                       &0.7      &\{0.3, 0.8, 0.5\} \\
        \hline
        Pick Carrot (real-world)      &Put the carrot into the plate.     &0.7      &\{0.8, 0, 0.5\} \\
        Pull Compartment (real-world) &Pull open the sliding compartment. &0.7      &\{0.8, 0, 0.5\} \\
    \end{tabular}
    }
  \end{center}
\end{table}

\begin{figure*}[t]
\centering
\includegraphics[width=\linewidth]{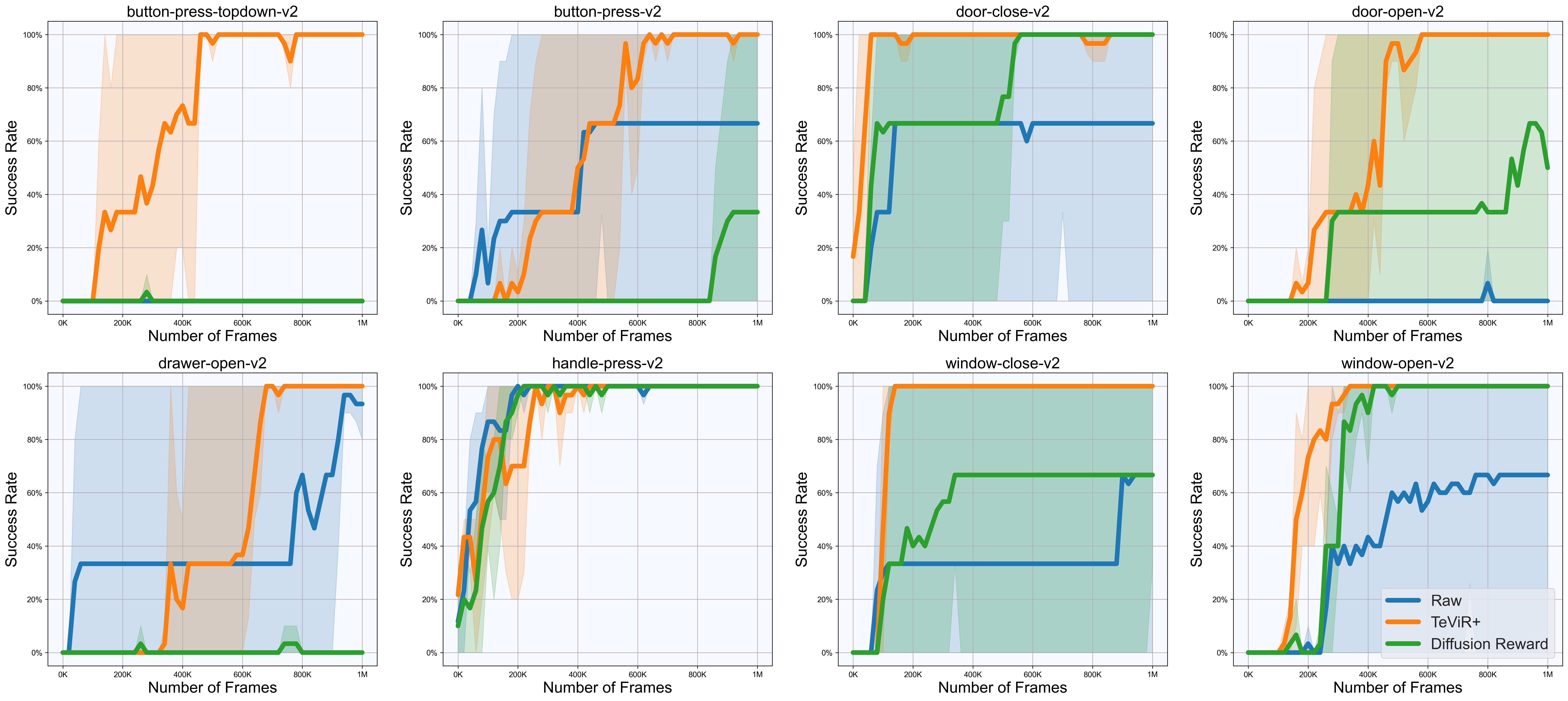}
\caption{Performances of the different reward engineering methods with the environmental sparse reward on 8 tasks from Meta-World benchmark. Raw Sparse Reward and Diffusion Reward are compared with TeViR+ with 3 random seeds within 1M training steps. }
\label{fig:env_result}
\end{figure*}

\begin{figure}[t]
\centering
\includegraphics[width=\columnwidth]{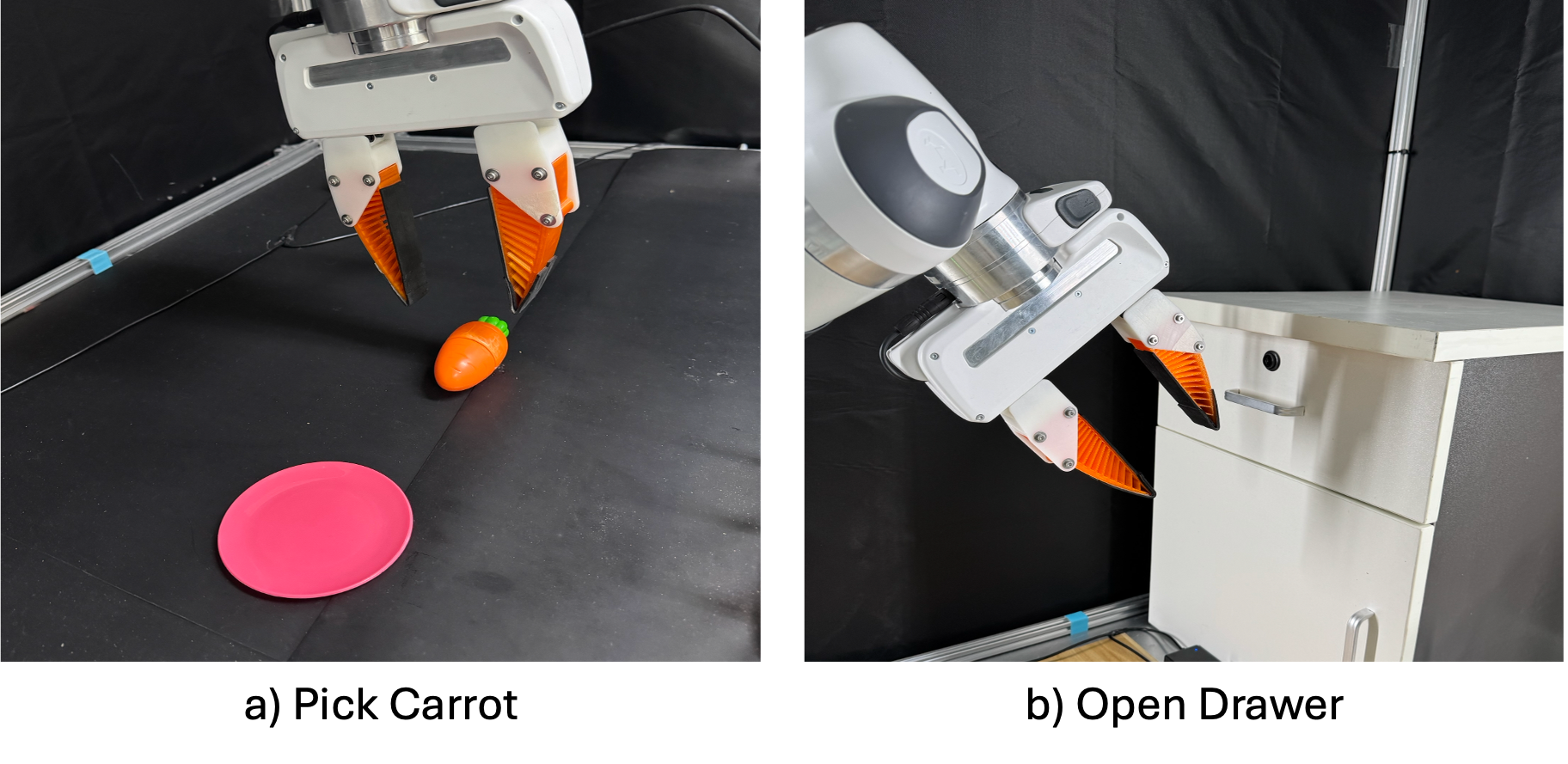}
\caption{Overview of all real-world experimental tasks, including a) Pick Carrot and b) Open Drawer.}
\label{fig:exp_real}
\end{figure}

\paragraph{Experimental Environments and Training Details}
We intend to demonstrate the effectiveness of TeViR on visual robotic manipulation tasks from Meta-World \cite{yu2019meta} with Sawyer arm, and real-world tasks with Franka Emika arm. To ensure task diversity and complexity when evaluating reward function, we choose 11 Meta-World manipulation tasks and 2 real-world tasks shown in Figure \ref{fig:exp_real}, including different skills such as reaching, pushing, and grasping. Each task is associated with $64 \times 64$-dimensional RGB images of multiple view and a binary ground truth environmental sparse reward. The sparse reward in real-world tasks is generated by a pre-trained ResNet-based classifier, which evaluates the completion of target task. These tasks are widely used in visual RL and are chosen to be diverse in objects and manipulating skills. The language description and hyperparameters for each task are shown in Table \ref{tab:param}.

For the Meta-World tasks, we collect 30 expert videos of each view for each via the scripted policy provided by the official repository\footnote{\url{https://github.com/Farama-Foundation/Metaworld}}. For the real-world tasks, we collect 30 expert videos of each view through human tele-operation. And we utilize the task description as shown in Table \ref{tab:param} to synthesize ten sentences with consistent semantic content to expand data diversity.

For the video generation, we used a fixed pre-trained CLIP-Text encoder (63M parameters) \cite{alec2021learnning} to encode language task descriptions into 512-dimension embedding. Then we follow AVDC \cite{chen2023learning} and use a similar network architecture and training setup which utilizes Video U-Net model with 3 residual blocks of 128 base channels and channel multiplier [1, 2, 3, 4, 5], attention resolutions [8, 16], attention head dimension 32. We use 1e-4 as the learning rate and Adam optimizer for model updating. The training is performed on an NVIDIA A6000 GPU with a batch size of 8, using the first 80\% of the video data, with the remaining 20\% used for evaluation. The video diffusion model is trained for 400K steps, taking approximately 20 hours to complete.

Note that for a fair comparison, all reward engineering methods tested are agnostic to the choice of downstream RL algorithm. For Meta-World tasks, we use DrQv2\footnote{\url{https://github.com/facebookresearch/drqv2}} \cite{denis2022mastering} as the backbone. And we use $action\_repeat = 3$ and $feature\_dim = 50$ for DrQv2 policy. For real-world tasks, we adopt HIL-SERL\footnote{\url{https://github.com/rail-berkeley/hil-serl}} \cite{luo2024precise}, incorporating human intervention during training. For all tasks, the discount is 0.99. We use 1e-4 as the learning rate and Adam optimizer for policy training, which is performed on an NVIDIA A6000 GPU with a batch size of 256.

\begin{figure*}[t]
\centering
\includegraphics[width=\linewidth]{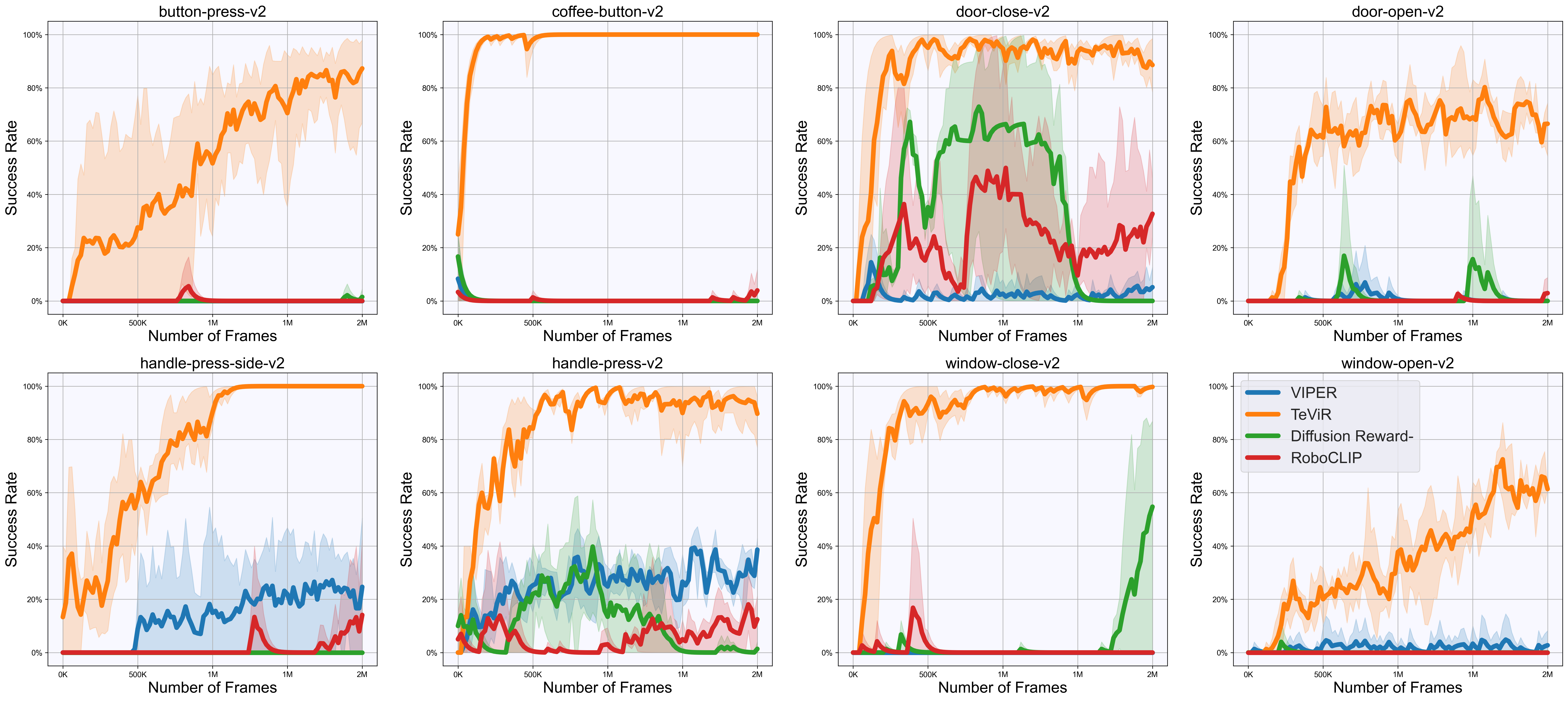}
\caption{Performances of the different reward engineering methods without the environmental feedback on 8 tasks from Meta-World benchmark. RoboCLIP, VIPER and Diffusion Reward- are compared with TeViR with 3 random seeds within 2M training steps.}
\label{fig:nenv_result}
\end{figure*}

\subsection{Learning with Ground Truth Sparse Reward}
\label{sec:exp_nenv}

Firstly, we aim to investigate whether our proposed framework, TeViR, can efficiently guide the agent to the desired behavior while tackling complex robotics tasks. We select 8 tasks of varying difficulty levels, requiring the agent to learn behaviors such as pushing, pulling, and grasping. In the context of learning with ground truth environmental sparse rewards, we choose TeViR+ using Equation \ref{eq:reward_env}, which incorporates the environmental reward as part of the reward and compares it against both raw sparse reward and the Diffusion Reward. And the interaction budget is set at 1 million for comparison.

The experiment results shown in Figure \ref{fig:env_result} indicate that relying solely on sparse environmental rewards facilitates progress in more straightforward tasks like 'door-close-v2', where the agent needs to slap and close the door. However, it encounters considerable difficulties with more complex tasks such as 'door-open-v2', where the agent must first grasp the doorknob and then pull the door open. Utilizing video prediction models, Diffusion Reward employs conditional entropy as a dense reward to guide the agent in learning the desired behavior. Success in more complex tasks, such as 'door-open-v2' and 'drawer-open-v2', demonstrates that Diffusion Reward can guide the agent by rewarding those expert-like trajectories. However, this video prediction model uses the diffusion model to predict only the following image based on historical images, limiting its ability to provide guidance for long-horizon policies. Despite the powerful representational capabilities of the diffusion model in capturing expert distributions, it still struggles to guide the agent efficiently, resulting in excessively long training steps.

In contrast, by utilizing the text-to-video diffusion model to generate the entire expert trajectory for guiding the agent, TeViR+ achieves superior policy guidance, as evidenced by the greater sample efficiency. This approach enables the agent to benefit from a more comprehensive and structured learning process, effectively reducing the training steps required to achieve the desired behavior. The extensive guidance provided by the text-to-video diffusion model allows the agent to better understand the long-term implications of its actions compared to models that predict only the next state. This holistic approach facilitates more efficient learning of complex tasks.

Also, the presence of ground truth environmental sparse rewards results in significant variance across the three methods in Figure \ref{fig:env_result}, primarily due to the binary success signal (0 or 1) provided. During the initial stages of training, agents employ random exploration strategies. Agents can rapidly learn the desired behaviors upon discovering a successful try and receiving the environmental sparse reward. In this context, the effectiveness of different methods in guiding the agent's policy learning becomes a critical factor for sample efficiency. In scenarios with only sparse rewards, the variance is at its highest, indicating that this method relies heavily on the agent's ability to explore desired behaviors through random exploration. Conversely, dense rewards such as the Diffusion Reward effectively guide the agent's policy learning, leading to faster learning rates and reduced variance. Furthermore, the TeViR+ exhibits even lower variance than the Diffusion Reward, suggesting that it has better guidance that enables agents to achieve rapid success on learning the desired policy.

\subsection{Learning without Ground Truth Sparse Reward}
\label{sec:exp_env}

\begin{figure}[t]
\centering
\includegraphics[width=\columnwidth]{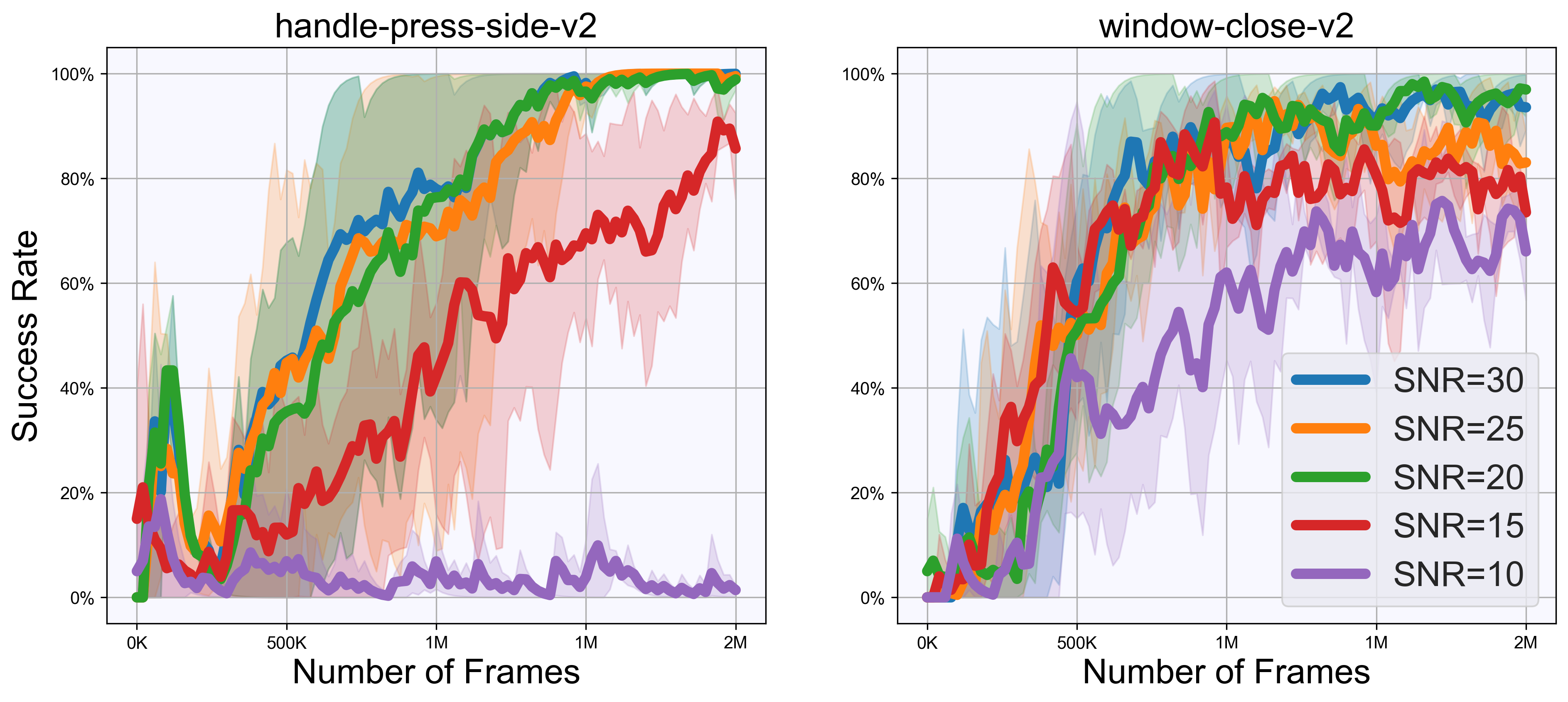}
\caption{Performances of TeViR with added noise on 2 selected tasks from Meta-World benchmark. Results are reported on 3 random seeds.}
\label{fig:noise}
\end{figure}

\begin{table}[t]
  \begin{center}
    \caption{Success rates with added noise.}
    \label{tab:noise}
    \scalebox{0.87}{
    \begin{tabular}{c|c|ccccc}
        \multirow{2}{*}{\bf Method} &\multirow{2}{*}{\bf Task} &\multicolumn{5}{c}{\bf SNR(dB)} \\ 
                               &                     &30       &25      &20      &15      &10 \\
        \hline
        \multirow{2}{*}{UniPi} &handle-press-side-v2 &90.0\%  &33.3\% &0.0\%  &0.0\%  &0.0\%\\
                               &window-close-v2      &83.3\%  &23.3\% &0.0\%  &0.0\%  &0.0\%\\
        \hline
        \multirow{2}{*}{TeViR} &handle-press-side-v2 &100.0\% &93.3\% &96.7\% &76.7\% &6.7\% \\
                               &window-close-v2      &93.3\%  &90.0\% &90.0\% &73.3\% &56.7\% \\
    \end{tabular}
    }
  \end{center}
\end{table}

To learn behaviors using our pre-trained text-to-video diffusion model and without utilizing any reward signals from the environment, we then provide the learning curves of success rates using Equation \ref{eq:reward_nenv} in TeViR. For our comparative analysis, we selected three methods that do not incorporate the ground truth environmental sparse rewards: RoboCLIP, VIPER, as well as the Diffusion Reward that removed $r^{spar}$ (labeled as Diffusion Reward-). This selection allows us to evaluate the impact of excluding traditional sparse reward mechanisms on the performance of each method. We also selected 8 tasks with varying difficulty levels, and the interaction budget is set at 2 million for comparison. All training curves are shown in Figure \ref{fig:nenv_result}.

The experimental results demonstrate that our method achieves outstanding performance across all tasks in the absence of the ground truth environmental sparse rewards. This finding underscores the robustness and effectiveness of our approach in complex robot manipulation tasks. As depicted in Figure \ref{fig:traj}, RoboCLIP, which generates a sparse reward by calculating the similarity between the image trajectory and the language description of the desired behavior, fails in almost all tasks. The agent can only get closer to its goal position, as language alone is insufficient to capture the intricacies of complex behaviors.

Meanwhile, both VIPER and Diffusion Reward- are capable of generating dense rewards to accelerate policy learning, yet they still suffer from low sample efficiency. Within the constraint of 2 million training steps, these methods also fail in most of the tasks. Notably, for the 'door-close-v2' task, the Diffusion Reward- experiences a policy collapse after initially achieving the desired behavior, indicating the instability of the conditional entropy reward without integrating the environment reward.

\begin{figure}[t]
\centering
\includegraphics[width=\columnwidth]{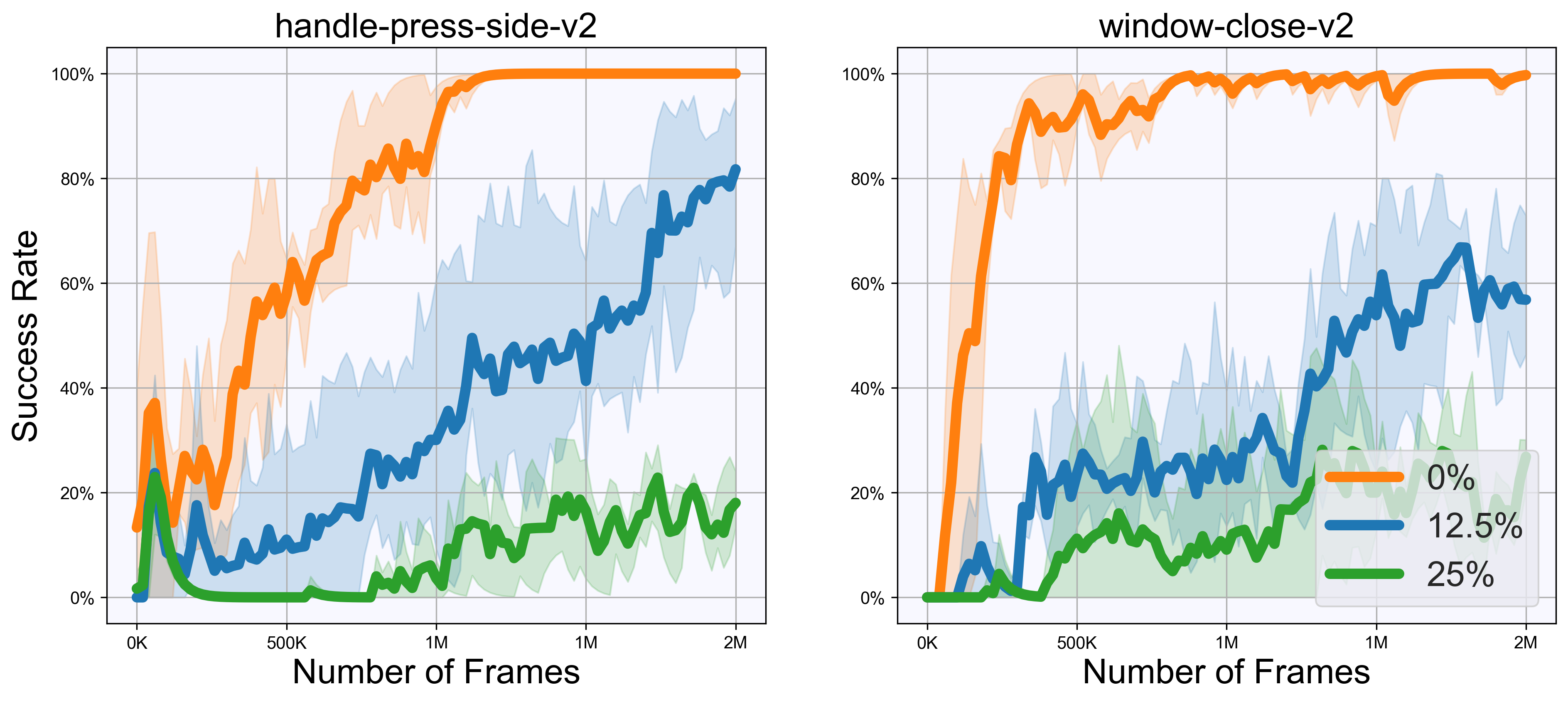}
\caption{Performances of TeViR with task-irrelevant and temporally disordered frames on 2 selected tasks from Meta-World benchmark. Results are reported on 3 random seeds.}
\label{fig:poor}
\end{figure}

\begin{table}[t]
  \begin{center}
    \caption{Success rates with erroneous frames.}
    \label{tab:error}
    \begin{tabular}{c|c|ccc}
        \multirow{2}{*}{\bf Method} &\multirow{2}{*}{\bf Task}          &\multicolumn{3}{c}{\bf Proportion of Error} \\ 
                                    &                                   &0\%      &12.5\%  &25\% \\
        \hline
        \multirow{2}{*}{UniPi} &handle-press-side-v2 &100.0\% &40.0\% &16.7\% \\
                               &window-close-v2      &96.7\%  &10.0\% &0.0\% \\
        \hline
        \multirow{2}{*}{TeViR} &handle-press-side-v2 &100.0\% &86.7\% &26.7\% \\
                               &window-close-v2      &96.7\%  &56.7\% &33.3\% \\
    \end{tabular}
  \end{center}
\end{table}

\subsection{Learning with Poorly Generated Video}
\label{sec:exp_unipi}
Maintaining temporal consistency across frames is a significant challenge for contemporary text-to-video generation methods. Existing models often produce videos with flickering and unnatural transitions as they struggle to preserve structural consistency throughout the video sequence \cite{yabo2023controlvideo}. In this section, we evaluate our method with UniPi to illustrate the performance of these two approaches under conditions of poor text-to-video generation quality.

Compared to the training outcomes of TeViR+ depicted in Figure \ref{fig:env_result}, the results for TeViR shown in Figure \ref{fig:nenv_result} demonstrate greater instability and an inability to achieve a 100\% success rate across multiple experimental settings (e.g., button-press-v2, window-close-v2). This instability is primarily due to the lack of ground truth sparse rewards from the environment, causing TeViR to converge to behaviors that merely approximate the generated expert trajectories. Consequently, the performance of TeViR heavily depends on these generated expert image sequences, which may not encompass all possible scenarios or offer optimal solutions for the present contexts.

To simulate poor-quality video generation, we employ two distinct methods. These evaluations aim to provide a comprehensive understanding of the robustness of our method and UniPi under adverse video generation conditions. All results for UniPi are implemented via open-loop control and tested on 30 independent experiment with random seeds. All results for TeViR are reported by 3 random seeds within 2M training steps.

\paragraph{Impact of Added Noise}
The first method involves directly adding noise to the generated videos. We investigate the performance of the two methods under varying signal-to-noise ratios (SNR). The results are shown in Table \ref{tab:noise}, and we provide training curves of TeViR in Figure \ref{fig:poor}.

\paragraph{Impact of Task-Irrelevant and Temporally Disordered Frames}
The second method introduces task-irrelevant or temporally disordered frames into the generated videos. We explore the final performance of both methods under a fixed proportion of erroneous frames. The results are shown in Table \ref{tab:error}, and we provide training curves of TeViR in Figure \ref{fig:poor}.

Based on the results in Table \ref{tab:noise} and Table \ref{tab:error}, it is clear that the presence of significant noise or incorrect frames in the generated video can significantly impact the performance of different methods. Specifically, UniPi's inverse dynamics model shows a marked sensitivity to these perturbations. High levels of noise and frame errors lead to notable inaccuracies in the inverse dynamics model's output, which in turn can result in policy failures. This sensitivity to noise can undermine the reliability of the UniPi in real-world applications where data imperfections are common.

Conversely, TeViR is based on reinforcement learning, enabling it to learn how to accomplish tasks from suboptimal states through exploration \cite{fu2025learning}, which contributes to its higher robustness compared to UniPi. Even when the Signal-to-Noise Ratio (SNR) is as low as 20 or the error rate is as high as 12.5\%, TeViR can still effectively guide the policy. Furthermore, when the SNR is reduced to 15 or higher error rates are introduced, TeViR still maintains a level of performance, albeit with some instability. This indicates that, even under more extreme conditions, TeViR retains its ability to adapt and continue learning, which showcases its adaptability and robustness in challenging real-world scenarios. However, under these conditions, the performance of TeViR is limited due to the challenges in accurately calculating the distance reward and progress reward, which are critical for guiding the agent's learning process.

\begin{figure}[t]
\centering
\includegraphics[width=\columnwidth]{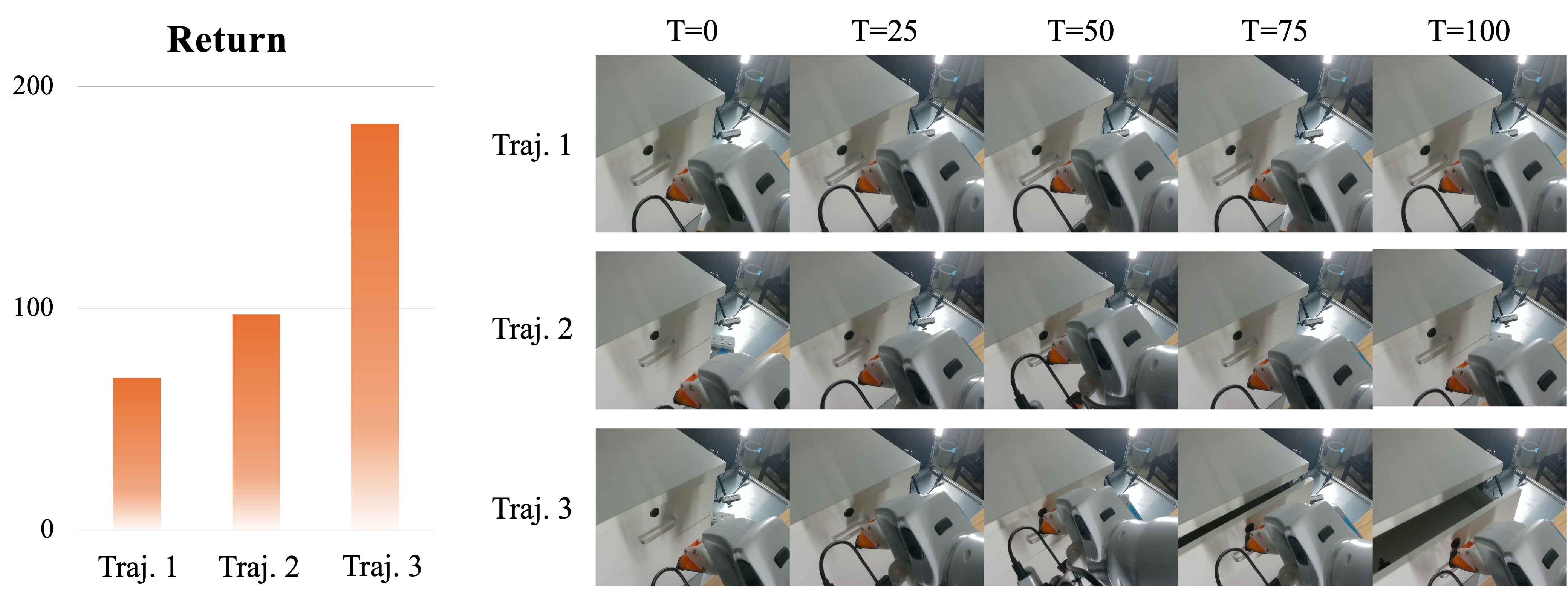}
\caption{Trajectory returns of TeViR on real-world "Open Drawer" task. Only Traj. 3 successfully completes the task.}
\label{fig:real_rwd}
\end{figure}

\begin{figure}[ht]
\centering
\includegraphics[width=\columnwidth]{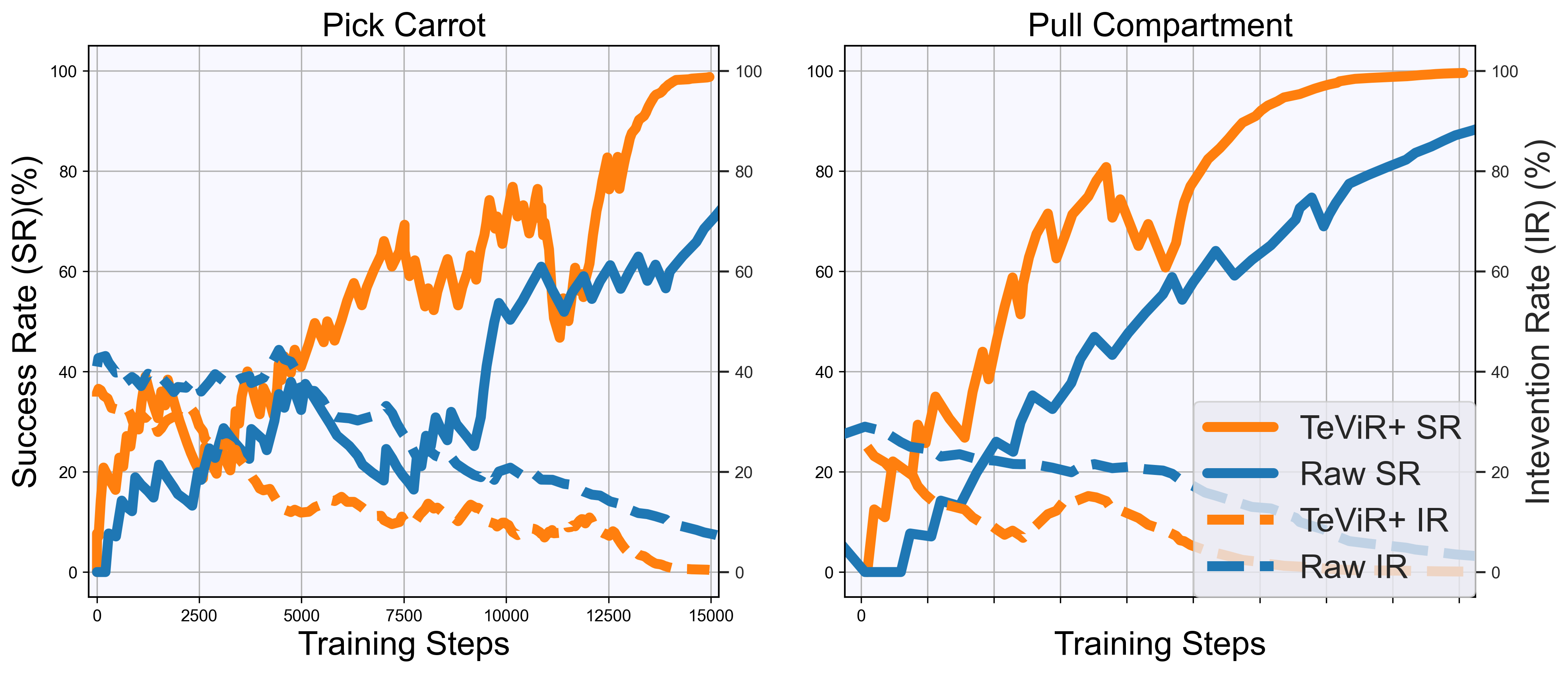}
\caption{Performances Raw sparse reward and TeViR+ on 2 real-world tasks. Raw Sparse Reward is compared with TeViR+ and all results are reported as a running average over 20 episode.}
\label{fig:real_d}
\end{figure}



\subsection{Real World Experiments}

To further assess the practical applicability of TeViR in real-world settings, where reward design remains a fundamental challenge, we extend our experiments to a physical robotic platform. Specifically, we evaluate RL training using TeViR+ against a baseline that relies solely on raw sparse rewards across two representative manipulation tasks, under otherwise identical training conditions.

As shown in Figure \ref{fig:real_rwd}, we first validate the effectiveness of TeViR rewards in a real-world "Open Drawer" task. In this setup, three trajectories reflect varying levels of task progress: Traj.1 exhibits only minor movement, Traj.2 fails to grasp the drawer handle, and only Traj.3 completes the task successfully. TeViR accurately distinguishes these differences, assigning rewards that reflect task success without relying on any hand-crafted metrics.

To fairly isolate the impact of TeViR+, we ensure a consistent level of human intervention across all settings. As shown in Figure \ref{fig:real_d}, TeViR+ significantly accelerates policy learning and leads to more stable and goal-consistent behaviors in real-world conditions. Notably, it achieves higher success rates within the same training steps, outperforming the sparse-reward baseline (e.g., 100\% vs. 65\% and 90\% in the two tasks). These results highlight TeViR’s potential to reduce the reliance on hand-crafted reward functions while preserving sample efficiency and robustness.

\begin{figure}[t]
\centering
\includegraphics[width=0.9\columnwidth]{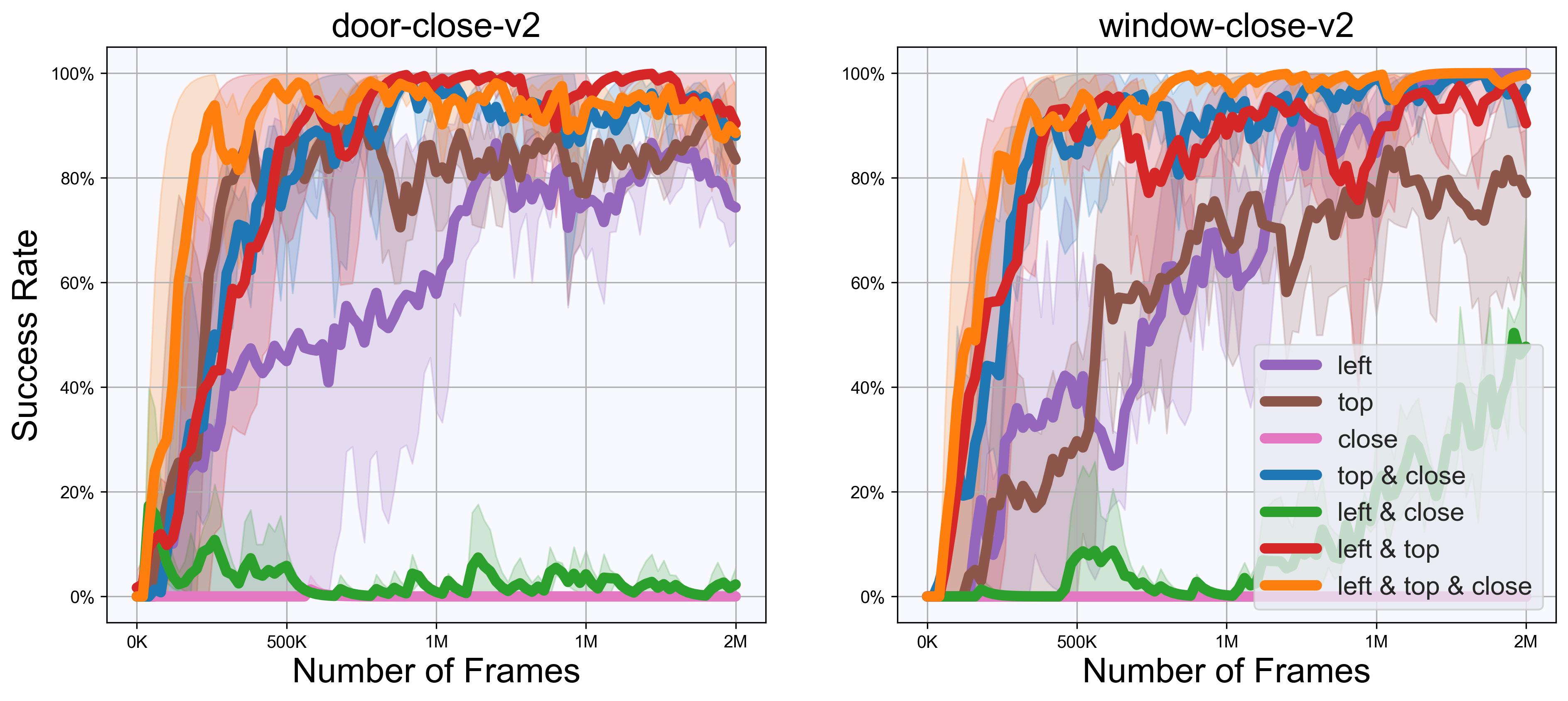}
\caption{Ablation studies for validating the necessity of multi-view observation on 2 selected tasks from Meta-World benchmark. Results are reported on 3 random seeds.}
\label{fig:exp_pose}
\end{figure}

\begin{figure}[t]
\centering
\includegraphics[width=\columnwidth]{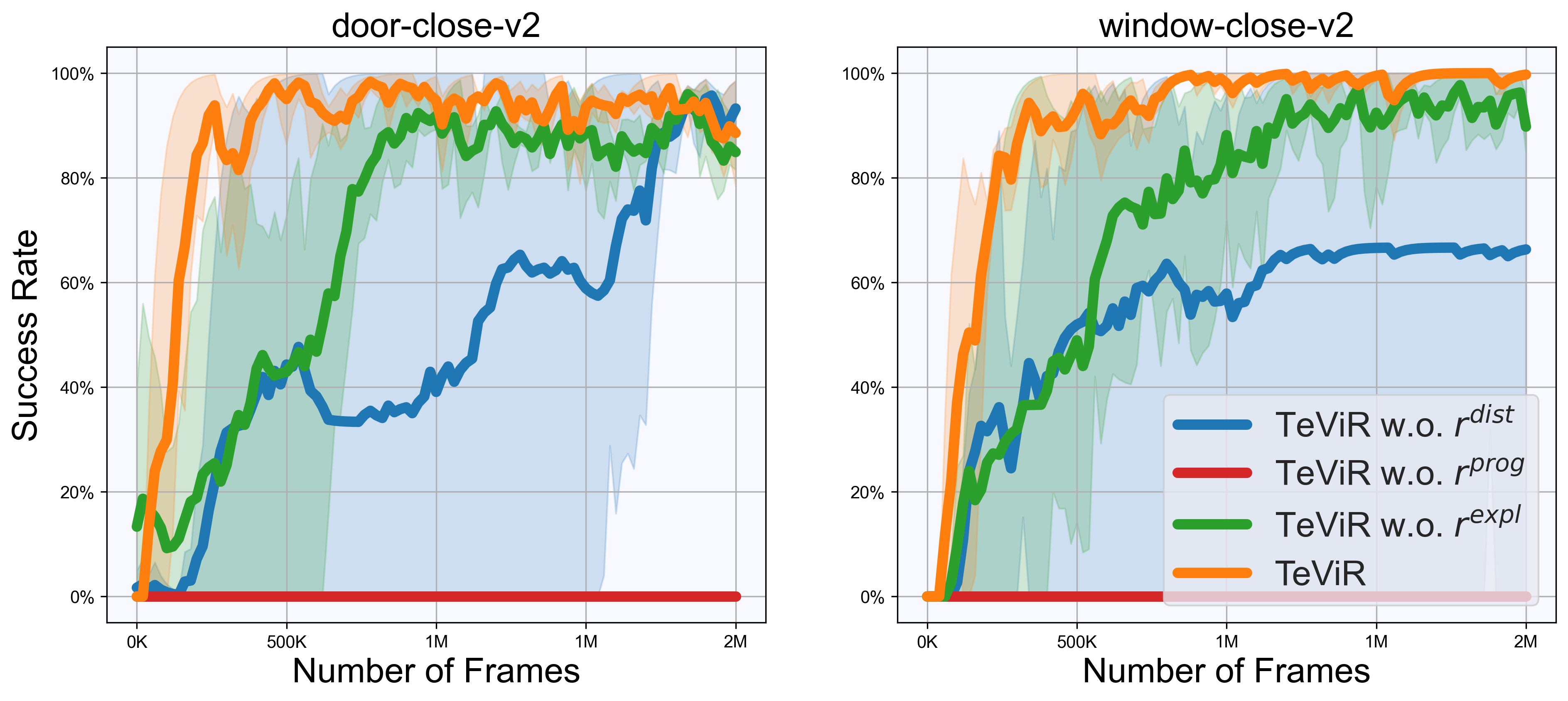}
\caption{Ablation studies for validating the necessity of each reward components of TeViR on 2 selected tasks from Meta-World benchmark. Results are reported on 3 random seeds.}
\label{fig:exp_abla}
\end{figure}

\subsection{Ablation Studies}
\label{sec:abla}
\subsubsection{Multi-view Observations}
\label{sec:exp_pose}
Selecting observations from multiple view is crucial for enhancing the environment representation, which directly impacts reward calculation, as evidenced in Figure \ref{fig:exp_pose}. In our Meta-World experiments, camera viewpoints are fixed, and we adjust a small number of hyperparameters, such as the weights assigned to each view, to align with the specific requirements of each task. The $close$ view primarily provides detailed information about the manipulated object, making it insufficient for guiding the agent effectively on its own. In contrast, training without the $close$ view may lose essential information about the manipulated object, which leads to unstable training, especially for those tasks that require careful manipulation on the object (e.g. 'window-close-v2'). On the other hand, the $left$ and $top$ view offer comprehensive information about the environment and the robot arm, leading to better performance when they are used. The weights assigned to each view can be adjusted to meet the specific requirements of different tasks and environments, as shown in Table \ref{tab:param}. For instance, the 'window-close-v2' task necessitates a parallel movement to close the window, thus requiring a higher weight on the $top$ view. Conversely, the 'drawer-open-v2' task demands precise manipulation of the handle, necessitating a higher weight on the $close$ view. Implementing multiple view can improve the accuracy of similarity calculations, offering more efficient and stable guidance, and thereby achieving better overall performance. While in real-world settings, we can place cameras to adequately capture the environment, the manipulated objects, and the robot arm’s state, while minimizing occlusions, without the need for additional hyperparameter tuning as shown in Figure \ref{fig:real_rwd}. This physical flexibility preserves generality across tasks without requiring manual adjustment of view weights, while still ensuring effective observations for reward computation and policy learning.

\subsubsection{Component Analysis of TeViR}
\label{sec:exp_rwd}
To investigate the impact of the different components $r^{dist}$, $r^{prog}$, and $r^{expl}$ that constitute TeViR on training performance, we conduct a series of controlled experiments as shown in Figure \ref{fig:exp_abla}. Each experiment isolate one of these components to evaluate its individual contribution to TeViR's overall performance. The distance reward $r^{dist}$ guides the agent toward the expected state by evaluating how close the agent is to the target. By providing feedback on similarity, this reward ensures that the agent's actions are directed towards reducing the gap between its current state and the desired state. The absence of $r^{dist}$ can lead to sub-optimal behaviors, where the agent may not prioritize finishing the task, resulting in inefficient learning.

On the other hand, the progress reward $r^{prog}$ assesses the task progress at the current state, giving the agent feedback on its advancement towards task completion. This reward is crucial for maintaining a sense of direction in learning procedure. Without $r^{prog}$, the agent lacks the necessary feedback of its progress, which often leads to learning failure as the agent may struggle to understand the significance of its actions in the context of the overall task.

Finally, the exploration reward $r^{expl}$ encourages the agent to explore the environment and discover new states and actions. This reward is vital for policy exploration, enabling the agent to gather diverse experiences and avoid being trapped in local optima. $r^{expl}$ helps in achieving efficient convergence to the desired behavior by ensuring that the agent explores sufficiently to find effective strategies for task completion. Without the exploration reward, the agent may not explore enough, leading to poor performance and slow convergence.

\section{Conclusion}
In this paper, we presented TeViR, a novel framework for reward engineering in RL, specifically designed to tackle the challenges of complex robotic manipulation tasks. By leveraging a pre-trained text-to-video diffusion model, TeViR generates dense rewards based on the similarity between predicted future states and current observations. Our extensive experiments across a variety of tasks demonstrated that TeViR outperforms existing methods, including those relying on sparse rewards and other advanced reward engineering methods. Notably, TeViR achieved better sample efficiency and performance, even in the absence of environmental sparse rewards. This highlights the robustness and effectiveness of our approach in complex environments. Additionally, TeViR's reliance on visual feedback and pre-trained video models offers a scalable and adaptable framework that can be applied to a wide range of tasks without the need for task-specific reward engineering. This represents a significant step forward in the development of general-purpose agents, paving the way for more efficient and effective reinforcement learning applications in robotics and beyond. 

Future work may explore several exciting directions to further enhance TeViR. One area of focus could be training video generation models with internet-scale data for a broader range of tasks, which enables TeViR to tackle more complex and realistic operation tasks, and drives advancements in the field of robotics by allowing for more sophisticated and real-world robotic applications. A more effective key frame selection or prediction mechanisms that are task-aware and semantically grounded is also important, to improve the fidelity and relevance of the generated videos and further reduce ambiguity during reward computation. Another promising direction is combining UniPi and TeViR to enhance sample efficiency and training robustness, enabling effective policy learning with a few samples. For instance, UniPi can be used to collect samples and warm up the policy, followed by TeViR for fine-tuning. This aims to achieve more efficient utilization of available samples and foster quicker adaptation to new tasks, which is crucial for advancing the capabilities of robotic systems in dynamic environments under resource-constrained condition. These enhancements aim to push the boundaries of what is achievable in reinforcement learning and robotic manipulation, driving forward the field towards more sophisticated and capable AI systems.

\section*{Acknowledgments}
This work is supported by the National Natural Science Foundation of China (NSFC) under Grants No. 62136008, No. 62103409, the Strategic Priority Research Program of Chinese Academy of Sciences (CAS) under Grant XDA27030400 and in part by the International Partnership Program of the Chinese Academy of Sciences under Grant 104GJHZ2022013GC.


\bibliographystyle{IEEEtran}

\vfill

\end{document}